\documentclass{article}
\usepackage{spconf,amsmath,graphicx}
\usepackage{bm}
\usepackage{enumitem}
\setlist{nosep, leftmargin=14pt}
\usepackage{cite}
\usepackage{mwe} % to get dummy images
\usepackage{xcolor}
\usepackage{colortbl}
\title{Reg-TTR, Test-Time Refinement for Fast, Robust and Accurate Image Registration}
\name{Lin Chen$^{1*}$, Yue He$^{1*}$\thanks{*\ These authors contributed equally to this work}, Fengting Zhang$^{1}$, Yaonan Wang$^{1}$, Fengming Lin$^{2}$, Xiang Chen$^{1,}$$^{\dagger}$\thanks{$^{\dagger}$Corresponding author: xiangc@hnu.edu.cn(Xiang Chen)},  Min Liu$^{1}$}
\address{$^{1}$ School of Artificial Intelligence and Robotics, Hunan University, China\\
$^{2}$ The University of Manchester, UK}

\begin{document}
%\ninept
%
\maketitle
\begin{abstract}
Traditional image registration methods are robust but slow due to their iterative nature. While deep learning has accelerated inference, it often struggles with domain shifts. Emerging registration foundation models offer a balance of speed and robustness, yet typically cannot match the peak accuracy of specialized models trained on specific datasets. To mitigate this limitation, we propose Reg-TTR, a test-time refinement framework that synergizes the complementary strengths of both deep learning and conventional registration techniques. By refining the predictions of pre-trained models at inference, our method delivers significantly improved registration accuracy at a modest computational cost, requiring only 21\% additional inference time (0.56s). We evaluate Reg-TTR on two distinct tasks and show that it achieves state-of-the-art (SOTA) performance while maintaining inference speeds close to previous deep learning methods. As foundation models continue to emerge, our framework offers an efficient strategy to narrow the performance gap between registration foundation models and SOTA methods trained on specialized datasets. The source code will be publicly available following the acceptance of this work. 

\end{abstract}
\section{Introduction}
\label{sec:intro}
%\begin{Introduction}

%traditional approach
%deep learning based
%foundation model
%our method

Image registration is a fundamental task in medical image analysis, with broad applications in surgical planning, navigation, motion tracking, and disease diagnosis \cite{chen2021deepSurvey}. Traditional methods \cite{vercauteren2009diffeomorphic} typically formulate registration as an iterative optimization problem, minimizing a composite objective function that combines a similarity metric with a smoothness regularization term. Although these conventional approaches demonstrate robustness and generalizability across diverse registration tasks (even on unseen data), their reliance on iterative optimization often results in substantial computational overhead and a time-consuming registration process.

%Deep learning-based methods \cite{balakrishnan2019voxelmorph}, by contrast, achieve rapid registration through a train-then-test paradigm. Once trained, the network performs registration in a single forward pass, typically completing in under one second. Training in a weakly-supervised manner, deep learning based appeaoches \cite{chen2024textscf} can significantly outperform traditional appeoaches, with the guidance of structure priors. While these approaches can match or even surpass the performance of conventional techniques, they face two fundamental challenges inherent to data-driven models: (1) dependence on substantial amounts of training data, and (2) limited generalization to out-of-distribution samples. As a result, retraining or fine-tuning is often required when deploying the model to a new registration task. And in those tasks with limited samples, traditiobnal appraoches can still outperform deeplearning based approaches.

Deep learning-based methods \cite{balakrishnan2019voxelmorph,chen2024textscf,zhang2024memwarp,wang_RDP,chen2021deepDiscontinuity1,eoir2026chen} achieve rapid registration through a train-then-test paradigm. Once trained, registration is accomplished via a single forward pass, typically in under one second. When trained in a weakly-supervised manner with structural priors \cite{chen2024textscf,zhang2024memwarp}, these approaches can significantly outperform traditional methods. Nevertheless, they face two fundamental challenges inherent to data-driven models: (1) dependence on substantial training data, and (2) limited generalization to out-of-distribution samples. Consequently, retraining or fine-tuning is typically required when deploying models to new registration tasks. In scenarios with limited data, traditional approaches may therefore still outperform their deep learning-based counterparts.

The emergence of foundation models \cite{uniGradICON2024MICCAI,11105528,10.1007/978-3-031-72390-2_62} offers strategies of both traditional and deep learning-based registration methods. By training on large-scale datasets, such a model can achieve robust performance across diverse registration tasks while maintaining inference speeds comparable to previous deep learning networks. However, despite these advantages in efficiency and generalizability, registration foundation models typically fall short of attaining optimal performance when compared to specialized, task-specific models on particular registration tasks \cite{uniGradICON2024MICCAI} .

To achieve optimal performance with registration foundation models, a natural strategy is to employ test-time optimization\cite{zhou2024test, Yuan_2024_CVPR, Karmanov_2024_CVPR, 10681158}, enhancing the model's outputs through pre- or post-processing techniques. The main current approaches for optimizing test time include adjusting all network parameters \cite{zhou2024test}, fine-tuning specific layers such as normalization layers \cite{Yuan_2024_CVPR},  adaptively optimizing features of the test images \cite{Karmanov_2024_CVPR}, and directly optimizing the parameters of the displacement field \cite{10681158}. However, this direction remains underexplored in medical image registration. For instance, uniGradICON \cite{uniGradICON2024MICCAI} introduced an instance optimization strategy that refines the network's parameters during inference to improve the accuracy of the registration. While effective, this method incurs a high computational overhead, limiting its practicality in real-world scenarios that demand fast registration and operate under constrained computational resources.
In this paper, we propose Reg-TTR, a novel medical image registration framework that integrates pre-trained registration networks with an efficient test-time refinement strategy to achieve robust, fast, and accurate registration across diverse clinical imaging scenarios. The proposed framework first leverages a registration foundation model to generate an initial deformation field, which is then iteratively refined by a dedicated test-time refinement (TTR) module using a novel composite loss function. This design preserves the essential topology of the initial displacement field, while enabling rapid convergence to an optimized solution with minimal computational overhead. We evaluated our method on two public datasets, including the Abdomen Computed Tomography (CT) dataset and the ACDC cardiac Magnetic Resonance(MR) dataset. In summary, the contributions of this work are threefold:
\begin{itemize}
\item We introduce Reg-TTR, a registration framework that effectively combines a foundation model with test-time refinement to achieve fast and robust performance across multiple registration tasks.
\item We propose a new composite loss function by combining the Structural Similarity (SSIM) loss with the commonly used Normalized Cross-Correlation (NCC) and smoothness regularization losses to enhance the robustness of image registration.
\item Our method matches or even surpasses the performance of specialized models on the ACDC and  Abdomen CT datasets, while maintaining computational efficiency on par with task-specific models. 
%Our method achieves state-of-the-art (SOTA) or comparable performance on the ACDC and an abdominal dataset, while maintaining computational efficiency comparable to task-specific deep learning methods.%
\end{itemize}

%\end{Introduction}
\begin{figure}[t]
\centering
\includegraphics[width=\columnwidth]{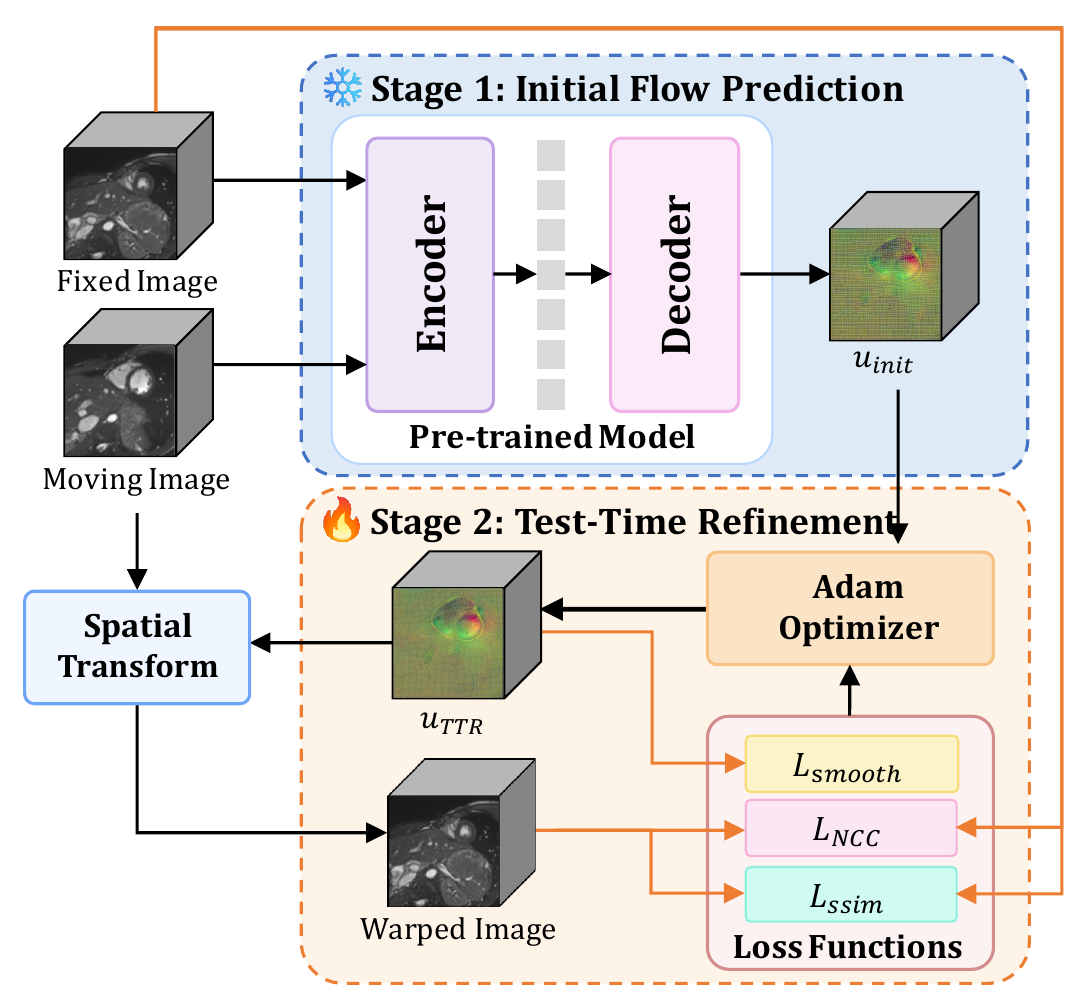}
\caption{The schema of our Reg-TTR. Reg-TTR framework operates in two stages: the first stage generates an initial deformation field using a pre-trained network, while the second stage further refines this field to achieve optimal registration performance. Reg-TTR effectively narrows the performance gap between general-purpose registration foundation models and data-specific models, thereby facilitating both rapid and accurate image registration.}
\label{fig:Reg-TTR} 
\end{figure}

\section{Method}

\subsection{Overall Framework}
% Image registration aims to find point correspondence between the moving image $I_M$ and fixed images $I_F$. The point correspondence is parameterized by a displacement field u(x), with the corresponding warp function as $\phi(x)$. The registration process is formulated as,
% \begin{equation}
% \phi(x) = x+u(x),
% \end{equation}
% \begin{equation}
% \hat{\textbf{u}} = \arg\min_{u} L(I_F,I_M \circ \textbf{u}),
% \end{equation}
% where L is the loss function to evalaute the different between the fixed image $I_F$ and warped moving image $\phi \circ u$. Once trained, deep learning network view this as a one forward inference step, directly predicting the $u$. However, our Reg-TTR framework view it as a `1+n' step process, including a forward inference to predict the initial deformation fiedls and n times of refinement step to further refine the initial deformation fields. 

Image registration establishes a dense correspondence between a fixed image $\bm{I}_F$ and a moving image $\bm{I}_M$. For deformable registration, this correspondence is parameterized by a deformation field  $\bm{u}(x)$, with the associated spatial transformation given by $\bm\phi(x) = x + \bm{u}(x)$. The registration objective is formulated as the optimization problem:
%\begin{equation}
%\hat{\mathbf{u}} = \arg\min_{\mathbf{u}} \mathcal{L}(I_F, I_M \circ \phi),
%\end{equation}%

\begin{equation}
\hat{\bm{u}} = \arg\min_{\bm{u}} \mathcal{L}( \bm{I}_F, \bm{I}_M \circ {\bm\phi}_{\bm{u}} ), \label{eq:registration}
\end{equation}
where $\mathcal{L}$ denotes the dissimilarity measure between the fixed image $\bm{I}_F$ and the warped moving image $\bm{I}_M \circ \bm\phi_{\bm{u}}$. Deep learning-based methods treat this as a single forward pass, directly predicting $\bm{u}$ after training. In contrast, our Reg-TTR framework conceptualizes registration as a `1+n' process: one initial deformation field prediction followed by $n$ iterative refinement steps to progressively enhance registration accuracy.

%The proposed Reg-TTR framework structure is illustrated in Fig. \ref{fig:reg_ttr}. This framework includes two stages, the first stage takes fixed and moving images as inputs,  obtaining an initial displacement field through a pre-trained model. The pre-traiend models can be either a pre-trained data-specific model or registration foundation model. Subsequently, the instance optimization module transforms the initial displacement field into a set of learnable parameters, deforming the moving image through spatial transformations. The optimization process employs Adam optimizer to minimize a hybrid function of similarity and smoothness regularization, with n steps. The optimization can be further accelerated via early stopping strategies. %, achieving an average optimization time of just 0.6 seconds. Combined with the average 0.2s duration for initial displacement field prediction, the complete registration workflow averages under one second. 

The proposed Reg-TTR framework, illustrated in Fig. \ref{fig:Reg-TTR}, consists of two sequential stages: the initial flow prediction and the TTR stages. In the first stage, fixed and moving images are fed into a pre-trained model, which may be a registration foundation model or a data-specific model (i.e., a model trained on data with the same distribution as the test data), to generate an initial deformation field $\bm{u_{init}}$. This field is converted to  a learnable parameter in the TTR stage. It refines the deformation over n iterations using the Adam optimizer to minimize a hybrid loss combining similarity and smoothness regularization. We adopt an early stopping mechanism based on loss monitoring: the optimization is terminated if the loss fails to improve for three consecutive iterations. 
%To improve computational efficiency, the optimization process can be accelerated using early stopping strategies.

\subsection{Test-time Refinement}
\label{subsec:ttr}
The second stage resembles a conventional iterative registration process as it refines the deformation field through multiple optimization steps. The key difference lies in initialization. While traditional methods start from an uninformed state (e.g., zero or random values), requiring many iterations and prolonged time, our Reg-TTR method begins with an informed initial estimate  $\bm{u_{init}}$. This strategic initialization enables the framework to achieve optimal registration performance within a significantly reduced number of iterations.
%The second stage resembles a conventional iterative registration process in that it refines the deformation field through multiple optimization steps. Unlike traditional approaches, however, which typically initialize the deformation field from zero or random values, requiring a large number of iterations to converge and resulting in prolonged registration times, our Reg-TTR method starts from an informed initial estimate $\bm{u_{init}}$. This strategic initialization enables the framework to achieve optimal registration performance within a significantly reduced number of iterations.

%To further refine the initial deformation fields $u_{init}$, we utilize Adam optimizer to optimze the deformation fields to the final deformation fields $u_{TTR}$ with hybrid loss function L. For each iteration, the moving image is warped by the deforamtion fields $u_{TTR}$ and then the hybrid function comprising similarity loss and smoothness regularization is computed based on the warped moving image and fixed image. Each voxel in the deofrmaiton fields is updated via the backpropogation.
To refine the initial deformation field $\bm{u_{init}}$, we employ the Adam optimizer to iteratively update the field toward the final deformation field $\bm{u_{TTR}}$, guided by a hybrid loss function $L$. At each iteration, the moving image is spatially transformed using the current estimate of $\bm{u_{TTR}}$, and the hybrid loss comprising a similarity term and a smoothness regularizer is evaluated between the warped moving image and the fixed image. The deformation field $\bm{u_{TTR}}$ is then updated via backpropagation, with adjustments applied at the voxel level to minimize the loss.

\subsection{Optimization Loss for TTR} 
In the TTR stage, we optimize the deformation field using a hybrid loss function comprising two similarity terms and a smoothness regularizer. The smoothness component, $L_{smooth}$, applies a diffusion regularizer to the spatial gradients of the displacement field $\bm{u_{TTR}}$  \cite{balakrishnan2019voxelmorph}. For similarity assessment, we adopt NCC, denoted as $L_{NCC}$\cite{chen2024textscf}, and additionally introduce a structural loss, $L_{ssim}$\cite{1284395}, aiming to further enhance the structural alignment between the warped moving image and the fixed image. The complete hybrid loss function is therefore defined as:
\begin{equation}
L = \lambda_1 L_{NCC} + \lambda_2 L_{ssim}
+ \lambda_3 L_{smooth}, 
\end{equation}
where the $\lambda_1$-$\lambda_3$ are the weights of each loss.

Considering that NCC primarily evaluates image similarity based on intensity information. To further improve registration performance, we introduce an SSIM loss to enhance the overall discriminative power of similarity assessment. The SSIM loss is specifically designed to address the limitations of the NCC term by explicitly penalizing structural differences between images. This results in more anatomically plausible registration outcomes and improves the robustness of detail alignment. The $L_{ssim}$ is formulated as,
\begin{equation}
\text{SSIM}(x, y) = \frac{(2\mu_x\mu_y + C_1)(2\sigma_{xy} + C_2)}{(\mu_x^2 + \mu_y^2 + C_1)(\sigma_x^2 + \sigma_y^2 + C_2)},
\end{equation}
where $\mu_x$ and $\mu_y$ are the local means of the images, $\sigma_x^2$ and $\sigma_y^2$ are the local variances of the images, $\sigma_{xy}$ is the local covariance between the images, $C_1 $ and $C_2 $ are stability constants.

\begin{table}[htbp]
\caption{Quantitative comparison on the Abdomen CT dataset. Statistically significant improvements in unsupervised registration accuracy are highlighted in bold. Symbols indicate direction: ↑ for higher is better, ↓ for lower is better.  uniGradICON* denotes utilizing uniGradICON’s own instance optimization.}
\label{tab:reg_ttr_abdomen}
\centering
\setlength{\tabcolsep}{0.pt} %
% \footnotesize  %  使用 footnotesize
% \resizebox{\textwidth}{!}{
%\vspace{4pt}
\small
\begin{tabular}{lccccc}
\hline
\rowcolor{gray!20}
Model &{Type}  & {Dice (\%) $\uparrow$} & {HD95 (mm) $\downarrow$} & {SDlogJ $\downarrow$} & Time (s) \\
\hline
Initial & - & 30.86 & 11.95  & 0.00 & - \\
\hline
VoxelMorph \cite{balakrishnan2019voxelmorph} & Semi & 47.05  & 23.08 & 0.13   & $< 1.0$\\ 
FourierNet \cite{jia2023fourier} & Semi & 42.80 & 22.95 & 0.13    & $< 1.0$\\
CorrMLP \cite{meng2024correlation} & Semi & 56.58  & 20.40  & 0.16   & $< 1.0$\\
\hline
VoxelMorph \cite{balakrishnan2019voxelmorph} & Un & 41.90  & 25.97 & 0.12   & $< 1.0$\\ 
FourierNet \cite{jia2023fourier} & Un & 41.83 & 25.25 & 0.11    & $< 1.0$\\
CorrMLP \cite{meng2024correlation} & Un & 51.01  & 22.80  & 0.13   & $< 1.0$\\
ConvexAdam \cite{10681158} & Un & 50.23  & 22.60  & 0.13   & 7.0 \\
uniGradICON \cite{uniGradICON2024MICCAI} & Un & 53.33  & 20.20 & 0.13  & 2.64\\
uniGradICON* \cite{uniGradICON2024MICCAI} & Un & 53.99  & 19.94 & 0.17  & 32.48 \\
\hline
\textbf{Reg-TTR(Ours)} & Un  & \textbf{56.81}  & 20.15  & 0.17  & 3.20 \\
w/o $L_{ssim}$  & Un & 56.45  & 20.10  & 0.15  & 3.18\\
\hline
\end{tabular}
% }
\end{table}

\section{Experiments }
\label{sec:experiments}
\subsection{Datasets and Implementation Details}
\textbf{Dataset and Image Preprocessing.} Reg-TTR was evaluated on two distinct registration tasks: intra-subject cardiac MR images registration and inter-subject abdominal CT registration. Cardiac data were obtained from the ACDC dataset \cite{bernard2018deep}, which includes 150 subjects divided into 85 for training, 15 for validation, and 50 for testing. For each subject, bidirectional registration was performed between end-diastole (ED) and end-systole (ES) phases, resulting in 100 test pairs. All cardiac images were resized to $128 \times 128 \times 16$ with a voxel spacing of $1.8 \times 1.8 \times 10$ mm\textsuperscript{3}. Abdominal CT scans are obtained from the Learn2Reg 2020 challenge \cite{xu2016evaluation}. The dataset was partitioned into 20 training, 3 validation, and 7 testing scans, generating 380 training pairs, 6 validation pairs, and 42 testing pairs. All Abdomen CT volumes were resampled to an isotropic 2 mm resolution and cropped to $192 \times 160 \times 256$. 

%Reg-TTR used Adam with learning rates of 0.1 for Abdomen CT and 0.025 for ACDC, and a ReduceLROnPlateau scheduler that halves the learning rate after two consecutive checks without loss decrease every two iterations.  All experiments were implemented in PyTorch and executed on a single NVIDIA RTX 3090 GPU. The hyperparameters were set to $\lambda_1=1$, $\lambda_2=2$, and $\lambda_3=1$. 

Our Reg-TTR utilized the Adam optimizer with a learning rate of 0.1 on the Abdomen CT dataset and 0.025 on the ACDC dataset. All experiments were implemented in PyTorch and executed on a single NVIDIA RTX 3090 GPU. The hyperparameters were set to $\lambda_1=1$, $\lambda_2=2$, and $\lambda_3=1$.  The source code of Reg-TTR is available on Github\footnotemark.
\footnotetext{https://github.com/VCL-HNU/Reg-TTR}

%We evaluate our Reg-TTR on two different registrtion tasks, intra-subject cardiac image registration and inter-subject abdomen image registration, using images from the ACDC  dataset \cite{bernard2018deep} and the abdominal CT dataset \cite{xu2016evaluation}. For images in ACDC dataset, 150 subjects werer split into training, validation and testing set, with 80, 20 and 50 subjects respectively. Bidirectional registration (ED-to-ES and ES-to-ED) was performed for each subject, yielding a total of 100 image test pairs. All images were preprocessed to a uniform size of $128\times128\times16$ with a voxel spacing of $1.8\times1.8\times10$ mm\textsuperscript{3} to ensure consistency. For the abdominal CT dataset sourced from the Learn2Reg 2020 challenge, we divided the dataset into three parts: 20 CT images for training, 3 for validation, and 7 for testing, which results in 380 ($20\times19$) training pairs, 6 ($3\times2$) validation pairs, and 42 ($7\times6$) testing pairs. All CT images were resampled to an isotropic resolution of 2 mm and resized to a spatial dimension of $192\times160\times256$.

\textbf{Evaluation Metrics.}  Following previous research \cite{balakrishnan2019voxelmorph}, we use the Dice Similarity Coefficient (Dice) to measure the overlap between segmented regions after registration. The 95\% Hausdorff Distance (HD95) is calculated to assess the maximum alignment error along anatomical boundaries. The Standard Deviation of the Logarithm of the Jacobian Determinant (SDlogJ) is used to evaluate deformation smoothness.

\textbf{Comparison Methods.} We compare our Reg-TTR framework with several unsupervised and semi-supervised baseline models, including VoxelMorph \cite{balakrishnan2019voxelmorph}, FourierNet \cite{jia2023fourier}, CorrMLP \cite{meng2024correlation}, and ConvexAdam \cite{10681158}.
To evaluate the generalization of the TTR optimization strategy, we compare the performance of the following models on the ACDC dataset before and after applying TTR optimization: VoxelMorph \cite{balakrishnan2019voxelmorph}, TransMorph \cite{chen2022transmorph},  RDP  \cite{wang_RDP}, MemWarp \cite{zhang2024memwarp}, and uniGradICON \cite{uniGradICON2024MICCAI}.
%To evaluate the generalization of our TTR optimization, we compared  five pre-trained models.  four data-specific models ( MemWarp \cite{zhang2024memwarp}, RDP  \cite{wang_RDP},  TransMorph\cite{chen2022transmorph}, and VoxelMorph \cite{balakrishnan2019voxelmorph}). Additionally, uniGradICON also incorporates an instance optimization strategy. However, uniGradICON optimizes the model's weight parameters, whereas our method directly optimizes the parameters of the displacement field. To rigorously evaluate the efficacy of our Reg-TTR, we also compare our Reg-TTR with the built-in optimization mechanism of uniGradICON.
%To assess our Reg-TTR, we selected five representative pre-trained models— MemWarp \cite{zhang2024memwarp}, RDP \cite{wang_RDP}, TransMorph \cite{chen2022transmorph}, VoxelMorph \cite{balakrishnan2019voxelmorph}and uniGradICON \cite{uniGradICON2024MICCAI}—using their initial displacement fields as inputs for test-time refinement. %

\begin{figure}[htbp] 
\centering
\includegraphics[width=\linewidth]{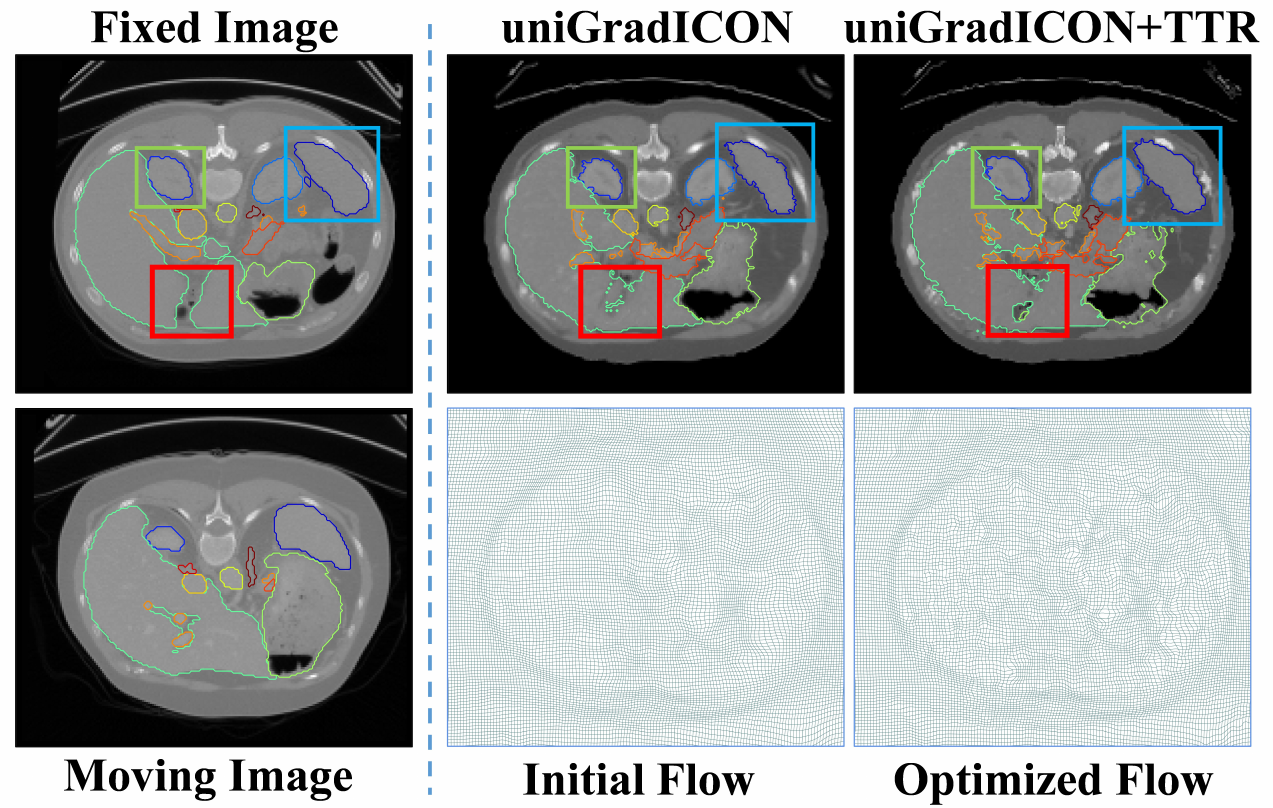}
\caption{Qualitative comparison on the Abdomen CT dataset. uniGradICON + TTR achieves a better structure consistency.}
\label{fig:visualization} 
\end{figure}

\subsection{Results and Analysis}

\textbf{Inter-subject Registration on Abdomen CT Dataset.} As shown in Table \ref{tab:reg_ttr_abdomen}, the results of the Abdomen CT dataset demonstrate that our Reg-TTR framework enables the foundation model uniGradICON to achieve a 3.48\% performance improvement, surpassing the semi-supervised CorrMLP \cite{meng2024correlation}. Compared to uniGradICON's built-in instance optimization (uniGradICON*), our Reg-TTR achieves a significant improvement in registration Dice, with significantly less time (3.2s vs 32.48s). As shown in Fig. \ref{fig:visualization}, Our proposed Reg-TTR framework delivers significantly better structure alignment for the uniGradICON foundation model while maintaining its smoothness, producing a final warped image that is notably closer to the fixed image.
%uniGradICON + TTR achieves a better structure consistency

\begin{figure}[!tbp]
\centering
\includegraphics[width=\linewidth]{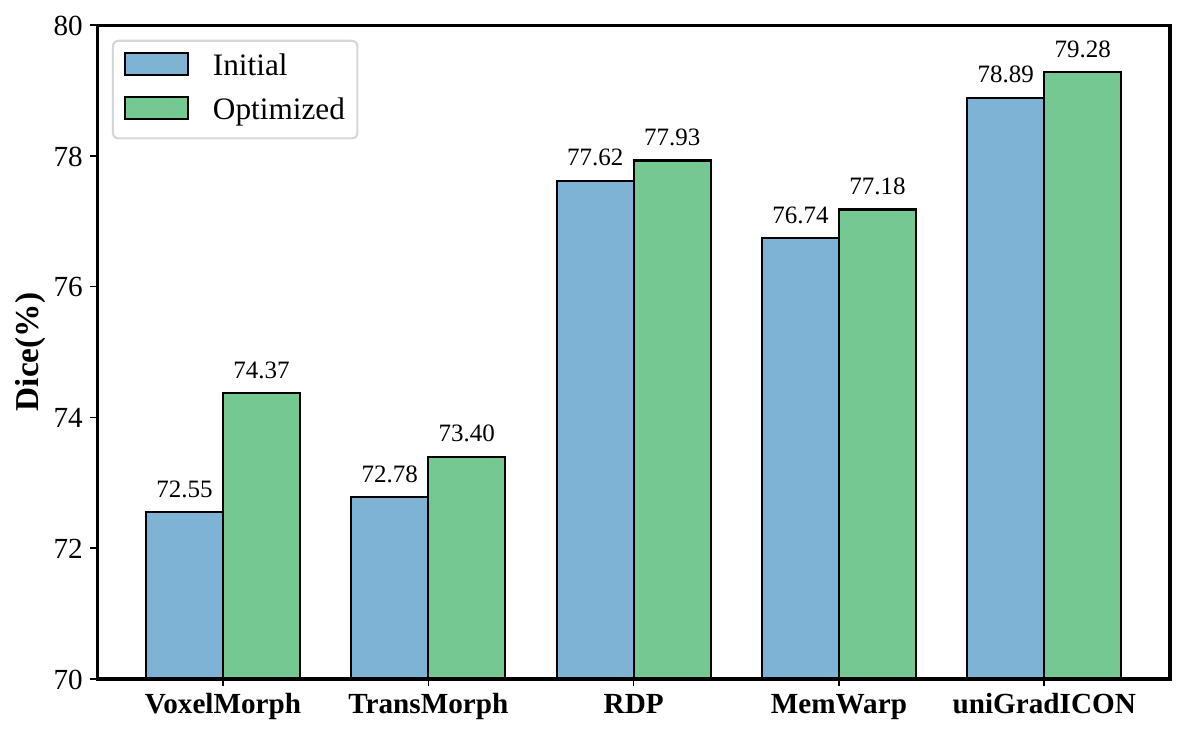}
\caption{Quantitative comparison of 5 pre-trained models on the ACDC dataset before (`initial') and after (`optimized') applying the TTR optimization.}
\label{fig:ACDC_dice} 
% \hline
% \end{tabular}
% }
\end{figure}

\textbf{Pre-trained Model Refinement on ACDC Dataset.} We further validate the Reg-TTR framework on uniGradICON and several other pre-trained models using the ACDC dataset, as illustrated in Fig. \ref{fig:ACDC_dice}. The results show that our TTR optimization consistently improves the Dice scores of all pre-trained models. 

\textbf{Analysis on SSIM Loss.} 
We further evaluate the contribution of the SSIM loss via an ablation study. We remove $L_{ssim}$ from the hybrid loss of Reg-TTR, denoted as `w/o $L_{ssim}$', and plot its results in Table \ref{tab:reg_ttr_abdomen}. The results demonstrate that $L_{ssim}$ can improve registration accuracy without sacrificing registration speed (3.20 s vs 3.18 s). In challenging scenarios like abdominal image registration, it can enhance the structure consistency and lead to robust improvements.

% \end{table*}
\begin{figure}[!t] 
\centering
\includegraphics[width=\linewidth]{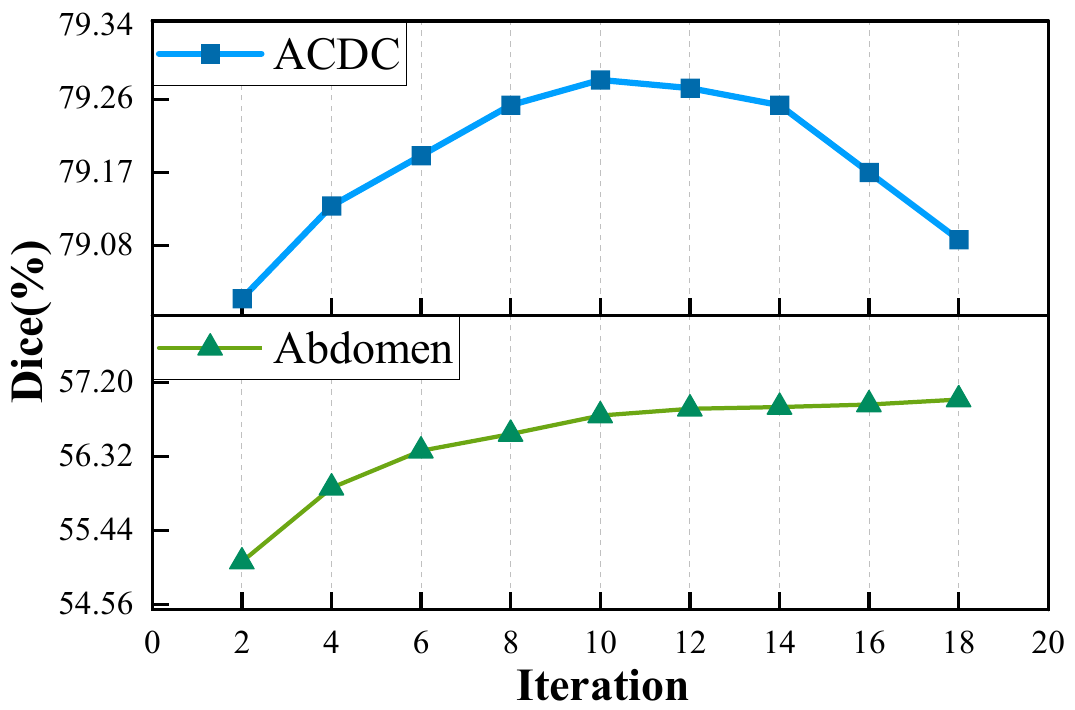}
\caption{Registration Dice scores of our Reg-TTR on ACDC and Abdomen CT datasets with increasing TTR iterations.}
\label{fig:acdc} 
\end{figure}

\textbf{Analysis on Iteration Number n.} 
The iteration number n balances the registration accuracy and time, which is a fundamental parameter in our Reg-TTR. We visualize the Dice improvement with increase of iteration number, as shown in Fig. \ref{fig:acdc}. It is observed that, due to the good start of initial deformation fields, our Reg-TTR can achieve the optimal registration performance within limited iterations. After that, the registration keeps waving or even drops with more optimization steps. According to Fig. \ref{fig:acdc}, we find that the Reg-TTR requires more iterations on the abdominal images compared with in the ACDC dataset, as more complex structures on the abdominal images are to be aligned with the abdominal images. However, n=10 would be  a good choice for both datasets.

% \begin{figure*}[t]  % 使用 figure*（带星号）跨双栏，[t] 表示顶部
% \centering
% \includegraphics[width=\textwidth]{img/可视化图v2(1).png}  % 使用整个页面宽度
% \caption{Qualitative comparison on the abdomen CT dataset}  % 添加标题
% \label{fig:visualization}  % 添加标签方便引用
% \end{figure*}

%
% \begin{figure}[!tbp] 
% \centering
% \includegraphics[width=\linewidth]{img/visualization.pdf}
% \caption{Qualitative comparison on the Abdomen CT dataset. uniGradICON + TTR achieves a better structure consistency.}
% \label{fig:visualization} 
% \end{figure}

\section{Conclusion}
%In this work, we propose the Reg-TTR framework, a novel test-time refinement paradigm that deeply leverages the complementary properties of deep learning and traditional registration methods. Specifically, a registration foundation model is utilized to predict an initial deofrmaiitn field. During inference, this initial field is defined as a set of optimizable parameters, optimized iteratively with a hybrid loss. This process effectively mitigates the domain shift between training and test data, enhancing registration accuracy while maintaining inference efficiency comparable to deep learning methods. Our experiments demonstrate that the Reg-TTR framework achieves SOTA registration performance on the ACDC dataset and abdominal CT dataset with only a few refinement steps, while maintaining registration times close to deep learning methods. Our Reg-TTR provides an effective solution for bridging the gap between the registration foundation model and data specific SOTA approaches, enabling fast, robust and accurate registration across various registration tasks.

In conclusion, we propose Reg-TTR, a test-time refinement framework that synergizes the robustness of iterative optimization with the efficiency of deep learning. The core of our approach lies in treating the foundation model's output as an optimizeable variable and refining it with a hybrid loss function during inference, which effectively combats domain shift and improves registration performance. Validated on both ACDC and Abdomen CT datasets, Reg-TTR achieves SOTA accuracy in merely a few refinement steps, with inference time remaining on par with standard deep learning methods. This work offers a practical and powerful strategy to narrow the performance gap between general registration foundation models and task-specific SOTA methods, paving the way for more versatile and efficient registration in clinical practice. Future work will extend the Reg-TTR framework to more datasets and multi-modal scenarios.

\section{Compliance with ethical standards}

This research utilized exclusively publicly available datasets. No new experiments involving human participants or animals were conducted by the authors. Therefore, ethical approval from an institutional review board was not required for this study. 

\section{Acknowledgments}
This work was supported by the National Natural Science Foundation of China (grant numbers U22B2050, 62425305, and 62503161), and in part by the science and technology innovation Program of Hunan Province under grant 2025RC3069, and in part by the Natural Science Foundation of Hunan Province under Grant 2025JJ60389.

\bibliographystyle{IEEEbib}  
\bibliography{refs}      
% \section{Referencing}
% \label{sec:ref}

% List and number all bibliographical references at the end of the
% paper.  The references can be numbered in alphabetic order or in order
% of appearance in the document.  When referring to them in the text,
% type the corresponding reference number in square brackets as shown at
% the end of this sentence \cite{C2}.

\end{document}